\documentclass[lettersize,journal]{IEEEtran}
\IEEEoverridecommandlockouts

\usepackage{makecell}
\usepackage{xcolor}
\usepackage{colortbl}
\usepackage{multirow}
\usepackage{amsmath,amssymb,amsfonts}
\usepackage{algorithmic}
\usepackage{algorithm}
\usepackage{graphicx}
\usepackage{subcaption}
\usepackage{enumitem}
\usepackage{marvosym}
\usepackage[hidelinks]{hyperref}

\usepackage{todonotes}
\usepackage[normalem]{ulem}
\usepackage{soul}

\paperwidth=\dimexpr \paperwidth + 6cm\relax 
\oddsidemargin=\dimexpr\oddsidemargin + 3cm\relax
\evensidemargin=\dimexpr\evensidemargin + 3cm\relax
\marginparwidth=\dimexpr \marginparwidth + 3cm\relax

\begin{document}

\title{DemoTuner: Automatic Performance Tuning for Database Management Systems Based on Demonstration Reinforcement Learning\\
}





\author{Hui~Dou,
        Lei~Jin,
        Yuxuan~Zhou,
         Jiang~He,
        Yiwen~Zhang*,
        and~Zibin Zheng,~\IEEEmembership{Fellow,~IEEE}
\IEEEcompsocitemizethanks{
\IEEEcompsocthanksitem Hui Dou, Lei Jin, Yuxuan Zhou, Jiang He  and Yiwen Zhang are with the School of Computer Science and Technology, Anhui University, Hefei, China.\protect\\
E-mail: $\{$douhui, zhangyiwen$\}$@ahu.edu.cn, \\ 
$\{$jinlei,  e22301255, e125221236$\}$@stu.ahu.edu.cn


\IEEEcompsocthanksitem Zibin Zheng is with the School of Software Engineering, Sun Yat-sen University, Zhuhai, China.\protect\\
E-mail: zhzibin@mail.sysu.edu.cn

\IEEEcompsocthanksitem Yiwen Zhang is the corresponding author.
}

}






\IEEEaftertitletext{\vspace{-0.8cm}}
\maketitle







\markboth{Journal of \LaTeX\ Class Files,~Vol.~14, No.~8, August~2021}%
{Shell \MakeLowercase{\textit{et al.}}: A Sample Article Using IEEEtran.cls for IEEE Journals}

\begin{abstract}
The performance of modern DBMSs such as MySQL and PostgreSQL heavily depends on the configuration of performance-critical knobs.
Manual tuning these knobs is laborious and inefficient due to the complex and high-dimensional nature of the configuration space.
Among the automated tuning methods, reinforcement learning (RL)-based methods have recently sought to improve the DBMS knobs tuning process from several different perspectives.
However, they still encounter challenges with slow convergence speed during offline training.
In this paper, we mainly focus on how to leverage the valuable tuning hints contained in various textual documents such as DBMS manuals and web forums to improve the offline training of RL-based methods.
To this end, we propose an efficient DBMS knobs tuning framework named DemoTuner via a novel LLM-assisted demonstration reinforcement learning method.
Specifically, to comprehensively and accurately mine tuning hints from documents, we design a structured chain of thought prompt to employ LLMs to conduct a condition-aware tuning hints extraction task.
To effectively integrate the mined tuning hints into RL agent training, we propose a hint-aware demonstration reinforcement learning algorithm HA-DDPGfD in DemoTuner.
As far as we know, DemoTuner is the first work to introduce the demonstration reinforcement learning algorithm for DBMS knobs tuning.
Experimental evaluations conducted on MySQL and PostgreSQL across various workloads demonstrate that DemoTuner achieves performance gains of up to 44.01\% for MySQL and 39.95\% for PostgreSQL over default configurations. Compared with three representative baseline methods including DB-BERT, GPTuner and CDBTune, DemoTuner is able to further reduce the execution
time by up to 10.03\%, while always consuming the least online tuning cost. 
Additionally, DemoTuner also exhibits superior adaptability to application scenarios with unknown workloads.
\end{abstract}





\section{Introduction}
Modern database management systems (DBMSs) such as MySQL and PostgreSQL provide many performance-critical knobs and their performance (throughput, latency, etc.) heavily relies on the configuration of these knobs. Considering the complex relationship between performance and knobs as well as the high dimensionality of configuration space, manually tuning the performance-critical knobs by database administrators (DBAs) is both labor-intensive and time-consuming. Consequently, how to automatically tune these knobs to improve the performance of DBMSs has attracted considerable interest from both academic and industrial sectors~\cite{duan2009tuning,van2017ottertune, cereda2021cgptuner,zhang2022towards, zhang2021restune,zhang2019CDBTune,xiong2017ath, bao2018autoconfig,li2019qtune,ge2021watuning,yang2024vdtuner} in the past few years.

Broadly speaking, state-of-the-art automated knobs tuning methods can be categorized into search-based, model-based and reinforcement learning (RL)-based. Specifically, search-based methods~\cite{duan2009tuning,cereda2021cgptuner,zhang2022towards, zhang2021restune,yang2024vdtuner} employ techniques such as Bayesian optimization~\cite{shahriari2015taking, hutter2011sequential, cowen2020hebo}  to iteratively explore the configuration space to identify optimal knob settings in an online manner. However, since these methods do not effectively utilize historical tuning experiences and lack awareness of the runtime environment, they usually require extensive time-consuming configuration evaluations and must restart from scratch when workload or hardware changes happen. Model-based methods~\cite{van2017ottertune, xiong2017ath, bao2018autoconfig} initially train an offline model to predict the relationship between knob configurations and performance, and then conduct online knobs tuning based solely on this predictive model. Consequently, these approaches usually require a large number of high-quality samples to develop an accurate performance prediction model. Moreover, the poor adaptability of conventional machine learning algorithms often necessitates model retraining when changes in workloads or hardware resources occur. In contrast, RL-based methods~\cite{zhang2019CDBTune,li2019qtune,ge2021watuning} learn policies that map current runtime states to an action (i.e., configuration of knobs) in a trial-and-error manner during the offline training phase. After training, the agent only needs a few sequential online tuning iterations to adapt to the practical application scenario so as to recommend satisfying knob configurations.

Recent RL-based methods have sought to improve the DBMS knobs tuning process from several perspectives~\cite{cai2022hunter,dou2024deepcat+,henderson2022blutune}. However, they still encounter challenges with slow convergence speed during offline training, which negatively impacts both the effectiveness (performance improvement) and efficiency (tuning costs) of the subsequent online tuning phase. On the one hand, without domain knowledge for knobs tuning under current DBMS runtime environment, the poor initial policy of a RL agent heavily impedes its offline training process. On the other hand, high-performing knob configurations are very sparse and distributed in the entire configuration space, potentially causing the RL agent to experience a prolonged period without significant positive feedback. Fortunately, current DBMS manuals, web forums and other documents offer a wealth of information about knobs tuning~\cite{trummer2023db}\cite{lao2025gptuner}. Either derived from domain knowledge or validated from practice, these valuable tuning hints can be utilized to enhance the offline training phase of RL-based knobs tuning methods, thereby improving the performance of their online tuning phase. Nevertheless, there are still two major pending challenges to achieve this promising objective:

\textbf{CH1: How to comprehensively and accurately  mine runtime-aware tuning hints from various documents?} Previous natural language processing (NLP)-enhanced DBMS knobs tuning methods such as DB-BERT~\cite{trummer2023db} only focus on textual information regarding the selection and configuration of performance-critical knobs contained in the related documents. However, both the importance of knobs and their optimal configuration are highly dependent on  the runtime environment such as workload characteristics, system states and resource usage, directly ignoring these runtime information may lead to the inappropriate usage of tuning hints. For instance, as shown in Table~\ref{tab:1}, text about the MySQL knob \textit{innodb\_log\_file\_size} recommends setting it to 4 GB for a \textit{write-heavy} workload rather than the common 512 MB.\ Additionally, some texts do not recommend any explicit knob values, instead, they only provide implicit tuning advice. For example, although the text about \textit{sort\_buffer\_size} does not include specific value, however, we can still infer that the knob value should be increased when workload is read-heavy or involves many sort operations. This direction guidance can help prevent the RL agent from taking incorrect actions. Regrettably, such valuable implicit tuning hints have been completely overlooked by previous studies~\cite{trummer2023db, lao2025gptuner}.

\setlength{\abovecaptionskip}{0.2cm}
\setlength{\belowcaptionskip}{0.2cm}
\setlength{\tabcolsep}{1.2mm}{
\begin{table}[t]
\caption{Examples of texts and the contained tuning hints.}
 \renewcommand{\arraystretch}{1.5}
    \centering
    \resizebox{\columnwidth}{!}{
        \begin{tabular}
      {p{0.09\columnwidth}!{\vrule width 1.5pt}%
       p{0.19\columnwidth}|p{0.19\columnwidth}!{\vrule width 1.5pt}%
       l}
             \Xhline{3\arrayrulewidth}
            \textbf{Text} &
            \multicolumn{2}{l@{\vrule width 1.5pt}}{\makecell[l]{Starting with innodb\_log\_file\_size\\= \textcolor{red}{512M} (giving 1GB of redo logs\\) should give you plenty of room \\for writes. If you know your appli-\\cation is \textcolor{blue}{write-intensive}, you can \\ start with innodb\_log\_file\_size = \\\textcolor{red}{4G}.}} &
            \makecell[l]{sort\_buffer\_size: For a \textcolor{blue}{read-}\\\textcolor{blue}{heavy} workload (again in \\which tables \textcolor{blue}{don’t have}\\\textcolor{blue}{indexes}), \textcolor{blue}{sort queries} \textcolor{red}{require}\\memory to complete \\a full sort.}
            \\
            \Xhline{1\arrayrulewidth}
            \textbf{Knob} &
            \multicolumn{2}{l@{\vrule width 1.5pt}}{\textit{innodb\_log\_file\_size}} &
            \textit{sort\_buffer\_size}
            \\ \hline
            \textbf{Value} &
            \makecell[l]{\textcolor{red}{512MB}} &
            \makecell[l]{\textcolor{red}{4GB}} &
            \textcolor{red}{increase}
            \\ \hline
            \textbf{Env} &
            \makecell[l]{\textcolor{blue}{/}} &
            \makecell[l]{\textcolor{blue}{write heavy}} &
            \makecell[l]{\textcolor{blue}{read heavy}\\\textcolor{blue}{no index}\\\textcolor{blue}{sort operation}}
            \\  \Xhline{3\arrayrulewidth}
        \end{tabular}
    }\label{tab:1}
\end{table}
}

\begin{figure}[t]
    \centering
        \includegraphics[width=0.95\columnwidth]{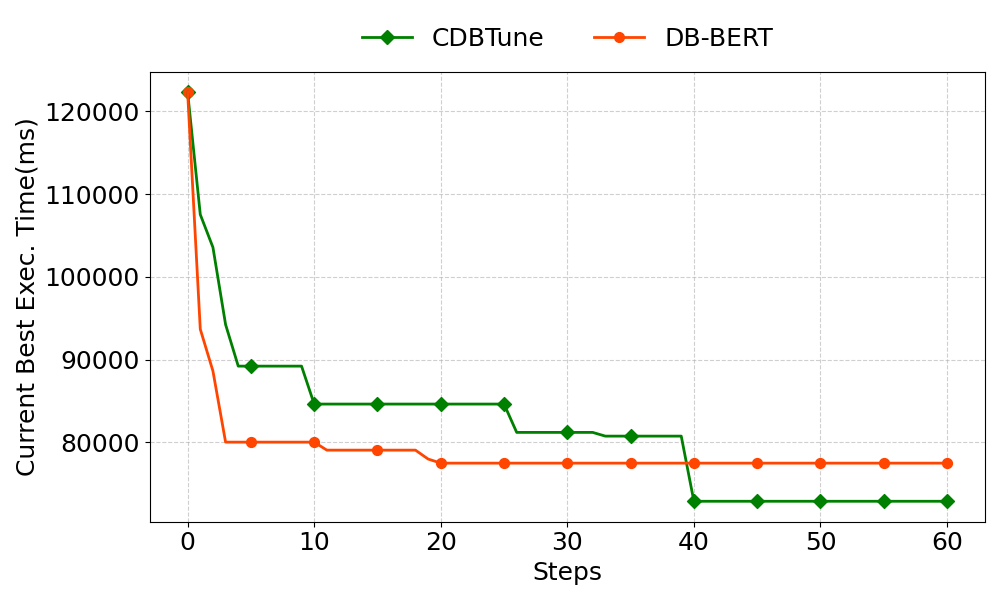}
        \caption{Current best performance  achieved by DB-BERT\cite{trummer2023db} and CDBTune\cite{zhang2019CDBTune} along with their RL agents training steps.}\label{fig:1}
\end{figure}

\textbf{CH2: How to effectively integrate the mined tuning hints into the RL agent training phase?} The standard RL training process involves an agent interacting with the environment to collect transitions through trial-and-error to determine optimal actions based on states, but it lacks support for the integration of domain knowledge. To our knowledge, DB-BERT~\cite{trummer2023db} is currently the  only  attempt to merge tuning hints with RL for  DBMS knobs tuning. This method utilizes a RL agent to translate tuning hints into knob values and prioritize hints according to runtime environmental feedback. However, in addition to the above mentioned shortcomings in tuning hints extraction, the knob configurations in DB-BERT are generated totally driven by tuning hints with a very limited range. On the one hand, the benefits of some
tuning hints tend to diminish along with knobs tuning iterations and may even deviate from the intended tuning target. On the other hand, transitions with real environmental feedback may provide more information than the static domain knowledge stored in tuning hints. As shown in Figure~\ref{fig:1}, compared with the pure DDPG-based method CDBTune~\cite{zhang2019CDBTune}, DB-BERT~\cite{trummer2023db} has an obviously larger convergence rate but gradually loses its competitiveness with ongoing training steps. Therefore, it is crucial to effectively incorporate the extracted tuning hints throughout the entire RL training phase to improve both the initial and long-term training performance of the tuning agent.


\setlength{\abovecaptionskip}{-0.0cm}
\setlength{\belowcaptionskip}{-0.4cm}

In this paper, we design and implement an efficient DBMS knobs tuning framework named DemoTuner via a novel
LLM-assisted demonstration reinforcement
learning method. To address $\S$CH1, we introduce a new application condition vector for tuning hints and design a structured CoT (chain of thought) prompt to employ LLMs to conduct the condition-aware tuning hints extraction task. With the advantages of LLMs
in semantic understanding and reasoning, DemoTuner is able
to mine both explicit and implicit tuning hints and correctly
summarize the corresponding application conditions. To address $\S$CH2, we propose a hint-aware demonstration reinforcement learning algorithm HA-DDPGfD in DemoTuner. Specifically, HA-DDPGfD consists of a pre-training phase and a fine-tuning phase. During the former phase, to mitigate the cold-start problem, HA-DDPGfD first generates demonstration experiences according to the mined tuning hints and then pre-trains the agent with these demonstrations. During the latter phase, we iteratively update the current priorities of tuning hints and provide a hint priority-driven prioritized experience replay mechanism called hpPER for agent training. In addition, we also develop a hint-guided reward shaping technique to further accelerate the convergence speed. With the trained agent, DemoTuner is able to execute online knobs
tuning under the user-specified constraint of iteration times.
To evaluate the effectiveness and efficiency of DemoTuner, a series of experiments are conducted on MySQL and PostgreSQL with different workloads generated from benchmark tools including YCSB and TPC-C. Experimental results show that the average best performance gain over default configurations achieved by DemoTuner reaches as much as 44.01\% for MySQL and 39.95\% for PostgreSQL, respectively. Specifically, compared with the representative baseline methods including DB-BERT~\cite{trummer2023db}, GPTuner~\cite{lao2025gptuner} and CDBTune~\cite{zhang2019CDBTune}, DemoTuner is able to further reduce the execution
time by respectively 10.03\%,  2.93\% and 7.85\% at most for MySQL workloads and 8.31\%, 5.41\% and 3.83\% at most for PostgreSQL workloads, while always consuming the least online tuning cost. Besides, DemoTuner also achieves the best adaptability to the simulated application scenarios with unknown workloads. The contributions of this paper are as following:
\begin{itemize}
\item To overcome the limitations of previous tuning hints extraction methods, we design a structured  prompt following the CoT framework to drive LLMs to mine both explicit and implicit tuning hints and
correctly summarize the corresponding application conditions.

\item We propose a hint-aware demonstration reinforcement learning algorithm HA-DDPGfD for agent training. As far as we know, DemoTuner is the first work to introduce the demonstration reinforcement learning algorithm for DBMS knobs tuning. In HA-DDPGfD, we develop a hint priority-aware PER mechanism and a hint-guided reward shaping technique to effectively employ the mined tuning hints to improve the agent training process.

\item We conduct a series of experiments on different DBMSs with various workloads. Experimental results successfully verify the advantages of DemoTuner over the baseline methods. The source code of DemoTuner is available at~\cite{DemoTuner2025Codes}.
\end{itemize}

The rest of this paper is organized as follows. Section~\ref{sec-2} introduces the preliminaries and related work. Section~\ref{sec-3} describes the system overview of DemoTuner. After that, Section~\ref{sec-4} and Section~\ref{sec-5} give a detailed description of DemoTuner. Section~\ref{sec-6} describes the experimental setups and Section~\ref{sec-7} presents the experimental results and analysis. Finally, Section~\ref{sec-8} concludes this paper.

\section{Preliminaries and Related Work}\label{sec-2}

\subsection{RL-based DBMS Knobs Tuning}
Knobs in DBMSs are closely related to their memory allocation, disk I/O, concurrency control and other factors. Knobs tuning means adjusting the values of knobs to improve the performance under specific execution environment. Reinforcement learning~\cite{sutton2018reinforcement} is a type of machine learning algorithms where an agent learns a policy to make decisions by performing actions in an environment to maximize the cumulative reward. In the context of DBMS knobs tuning, RL-based methods~\cite{zhang2019CDBTune,li2019qtune,ge2021watuning,li2023auto}  treat it as a trial-and-error process based on system feedback and exhibit a degree of adaptability to unknown runtime environments. Although recent studies~\cite{cai2022hunter,dou2024deepcat+,henderson2022blutune,mai2024ctuner,wang2021udo,zhang2023unified} have enhanced the RL-based methods from  various angles, training reinforcement learning models is still significant costly, especially in the field of knobs tuning, where each configuration evaluation for transition collection may require significant time consumption. The offline agent training usually suffers from the poor initial policy since the high-performing configurations are very sparse in the configuration space.

\subsection{Tuning Hints Extraction}

Manuals, web forums and other texts often offer a wealth of domain knowledge about knob settings for DBMSs. DB-BERT~\cite{trummer2023db} leverages BERT to mine these tuning hints in the form of $<$\emph{knob}, \emph{rec\_value}$>$, where \emph{rec\_value} is the recommended value for the corresponding $knob$. However, as shown in Table~\ref{tab:1}, since some texts only provide implicit tuning advice such as increasing and decreasing the knob, the extracted tuning hints should contain these tuning direction information. Besides, we should also pay attention to their application condition if we want accurately using the extracted tuning hints. Towards this target, we extend the definition of tuning hint as a recommended adjustment action for a knob under a specific runtime environment (i.e., application condition). Specifically, a condition-aware tuning hint can be described as a triple $<$\emph{knob}, \emph{rec\_action}, \emph{condition\_vector}$>$, where \emph{condition\_vector} is a vector describes the applicable runtime environment of \emph{rec\_action}. We will give a detailed description about the definition of \emph{condition\_vector} and how to extract condition-aware tuning hints in Section~\ref{sec-4}.

\setlength{\abovecaptionskip}{-0.0cm}
\setlength{\belowcaptionskip}{-0.3cm}
\begin{figure*}[t] 
\centering 
\includegraphics[width=1\textwidth]{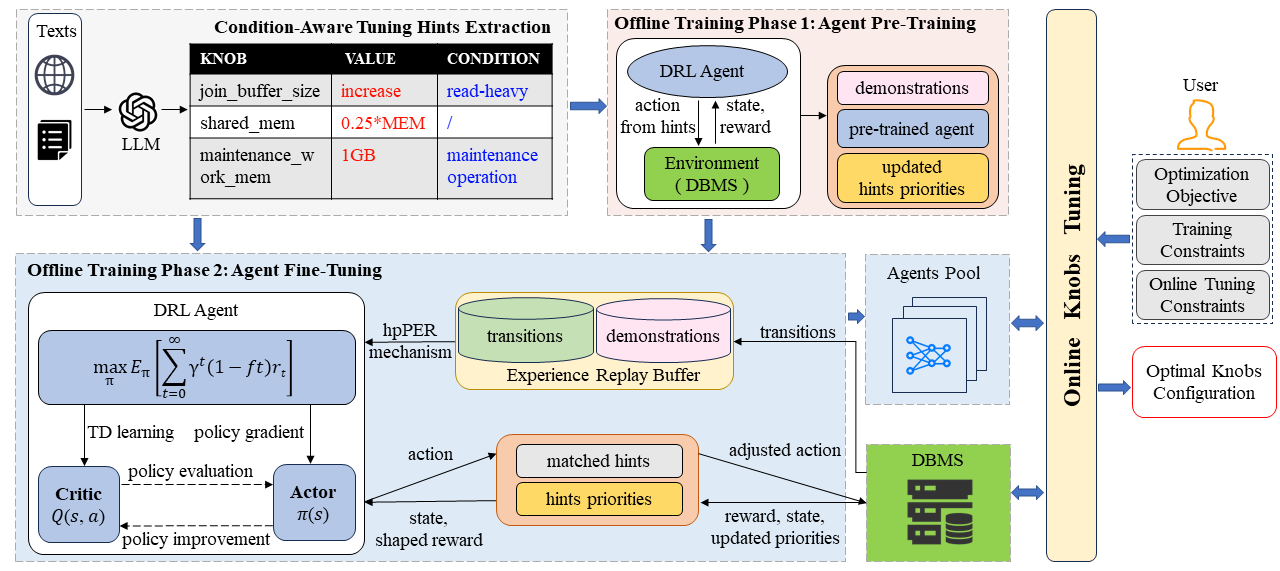} 
\caption{The system overview of DemoTuner.}\label{Fig.overview1} 
\end{figure*}

\subsection{LLM-Assisted and LLM-Driven Knobs Tuning}

As the pioneering work of NLP-enhanced DBMS tuning methods, DB-BERT~\cite{trummer2023db} first utilizes BERT to extract the tuning hints from texts, and then leverages RL to deal with these tuning hints to generate knob values in an iterative manner. Although DB-BERT can incorporate domain knowledge into the knobs tuning task, however, it totally neglects the application condition of tuning hints, which may mislead the tuning agent. In addition, it is only able to explore a limited
range around the recommended knob values by tuning
hints, which may miss the opportunities to further improve the
performance. Finally, as shown in Figure~\ref{fig:1}, totally driven by tuning hints may lose competitiveness along with training steps since the tuning hints' utilities in performance improvement are not static. Considering the advantages of large language models (LLMs) in semantic understanding, reasoning and domain knowledge, several recent works begin to use LLMs for DBMS knobs tuning. We broadly categorize them into LLM-assisted~\cite{lao2025gptuner, huang2024llmtune} and LLM-driven~\cite{fan2024latuner, li2024large,thakkar2025elmo,giannakouris2025lambda}. The LLM-assisted methods usually employ a LLM to extract or output tuning hints and then conduct a traditional knobs tuning procedure. For instance, GPTuner~\cite{lao2025gptuner} uses GPT-4 to select performance-critical knobs and extract the suggested value ranges, and then conduct a Bayesian optimization (BO)-based knobs tuning process. However, tuning hints extracted by the LLMs are not always reliable~\cite{liang2025same}, relying solely on GPT's judgments does not ensure the alignment and accuracy of domain knowledge with the current environment since BO
inherently lacks the ability to interact with runtime environment. We also employ LLMs to assist the tuning hints extraction task in DemoTuner. On the other hand, LLM-driven methods design prompts for the whole knobs tuning workflow including knobs selection, knobs range determination and knobs configuration. While promising, relying solely on LLMs to conduct knobs tuning is still full of challenges at the present stage considering the hallucination and domain adaptation issues.

\section{Overview of DemoTuner}\label{sec-3}
In this paper, we propose DemoTuner, an efficient DBMS knobs tuning framework via LLM-assisted demonstration reinforcement learning. As shown in Figure~\ref{Fig.overview1}, DemoTuner consists of three major modules including the condition-aware tuning hints extraction module, the offline agent training module and the online knobs tuning module. The core innovation of DemoTuner lies in  employing a novel demonstration reinforcement learning-based method to efficiently make use of the domain knowledge to accelerate the offline agent training. We give a brief introduction of each module below.

\textbf{Condition-Aware Tuning Hints Extraction:} This module is responsible for mining
tuning hints from various documents comprehensively and accurately. To record the application condition of tuning hints, we introduce a new condition vector to  describe the runtime environment information including workload characteristic, system state and resource usage. After that, we design a structured prompt following the chain of thought framework to guide the LLMs to conduct the condition-aware tuning hints extraction task. The detailed descriptions of this module can be found in Section~\ref{sec-4}.

\textbf{Offline Agent Training:} This module is responsible for utilizing the mined tuning hints to achieve an efficient offline agent training process so as to improve the effectiveness and efficiency
of the subsequent online tuning phase.
Specifically, the proposed hint-aware demonstration reinforcement learning algorithm HA-DDPGfD consists of a pre-training phase and a fine-tuning phase. During the agent pre-training phase, HA-DDPGfD  generates demonstrations based on the extracted tuning hints and then pre-trains the agent. Since the benefits of tuning hints are varying with the training steps, during the agent fine-tuning phase, HA-DDPGfD dynamically adjusts the priorities of tuning hints and trains the agent with a hint priority-driven prioritized experience replay mechanism hpPER.\ Besides, a hint-guided reward shaping technique is also developed in HA-DDPGfD to further improve the speed of convergence. Details can be found in Section~\ref{sec-5}.

\textbf{Online Knobs Tuning:}
With the trained agent (stored in the Agents Pool with its applicable workload characteristics) during the offline phase, DemoTuner is able to execute the online knobs tuning tasks for the target DBMS under the user-specified constraints such as total configuration evaluation times. After that, the best knobs configuration ever explored and the achieved performance improvement is output to users. 

\section{Condition-Aware Tuning Hints Extraction}\label{sec-4}

\subsection{Defining the Condition Vector for Tuning Hints}\label{sec:4-1}
To extract domain knowledge in the form of tuning hints, we first systematically collect source texts from various documents such as web forums, blog posts, official manuals and configuration files. The collection process involved using a set of search keywords such as \emph{PostgreSQL performance tuning} and \emph{MySQL configuration tuning}.
To extract relevant content efficiently from these source documents, we also employ  specially designed regular expressions for data retrieval. Notably, currently we mainly focus on how to accurately and comprehensively mine tuning hints with LLMs and leave a more comprehensive solution for source texts collection as future work. In practice, the importance of knobs is highly dependent on the runtime environment (we call condition in this paper) such as workload characteristics and resource usage. Hence, we distinguish their conditions from the following three perspectives:
    \begin{itemize}[leftmargin=*]
\item \textbf{Workload Characteristics}:
This condition perspective helps describe whether a knobs tuning action is applicable under OLTP  (Online Transaction Processing) or OLAP  (Online Analytical Processing) workloads. OLTP workloads primarily involve write-heavy queries such as inserts and updates, whereas OLAP workloads consists of complex read-intensive queries. For OLAP workloads, we further subclassify them based on specific criteria including data volume, operational complexity, and types of operation such as WAL (write-ahead logging), joins, sorting, and hashing. The workload type can be identified through status indicators provided by the DBMS.

\item \textbf{System State}:
This perspective determines the suitability of configuration actions for various system states, such as \emph{dirty data in kernel page}, \emph{caching issues}, and \emph{low buffer ratio}. We leverage Prometheus to monitor the operating system and DBMS runtime metrics and then identify current system states according to some pre-defined rules. For instance, a \emph{low buffer ratio} system state usually indicates insufficient buffer pool capacity, can be diagnosed by comparing the metrics  \emph{innodb\_buffer\_pool\_read\_requests} with \emph{innodb\_buffer\_pool\_reads}.

\item \textbf{Resource Usage}:
This perspective describes which configuration actions are applicable based on resource usage. For instance, increasing the value of PostgreSQL knob \emph{shared\_buffers} can benefit the performance only when the server memory is sufficient, otherwise, it may lead to performance degradation or system crashes. We use the operating system runtime metrics to assess resource usages including CPU, memory, disk, etc.

\end{itemize}

\subsection{LLM-based Tuning Hints Extraction}
\label{LLM-based}
\begin{figure}[t] 
    \centering 
    \includegraphics[width=0.5\textwidth]{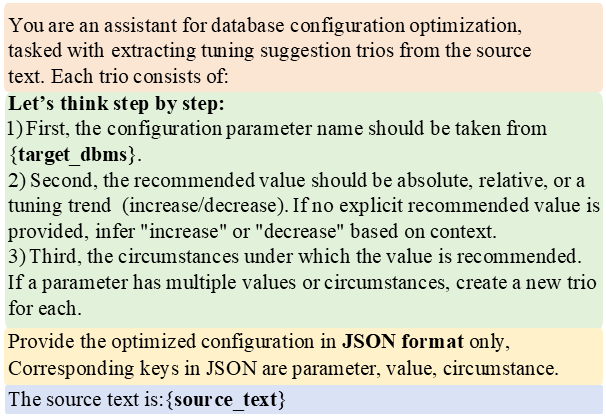} 
    \caption{Prompt construction for condition-aware tuning hints extraction.}\label{Fig.prompt} 
    \end{figure}

To effectively mine tuning hints described as $<$\emph{knob}, \emph{rec\_action},
  \emph{condition\_vector}$>$ from the collected source texts, we choose to utilize GPT-3.5-Turbo~\cite{gpt2025} after evaluating a range of advanced large language models (LLMs) considering the tradeoff between costs and practical effectiveness. How to select the best LLM is out of the scope of this paper.
Figure~\ref{Fig.prompt} illustrates the prompt structure, which includes four critical elements: the model's role and expertise, the input source text, the output format, and a task description following the chain of
thought (CoT) framework~\cite{wei2022chain}. Specifically, the CoT process can be divided into three steps. The first involves the extraction of knobs specific to the target DBMSs such as MySQL or PostgreSQL.\ The second focuses on extracting recommended values of knobs. In cases where no specific value is provided, the model is required to infer whether the knob's value should be increased or decreased. The third targets to extract the corresponding circumstances, which can be utilized to identify the application condition vector of this tuning hint. Incorporating the phrase \emph{``Let's think step by step''} to split up the task helps guide GPT's inference process. Besides, placing constraints directly within each step improves GPT's comprehension and execution by keeping the task and constraints closely aligned. These prompts sharpen  GPT's focus on DBMS knobs tuning and structure the task into clear steps, thus enhancing the extraction accuracy~\cite{kojima2022LLMZeroShot}. Notably, GPT's responses sometimes diverge from the prescribed format and exhibit variability. In such cases, we employ GPT to further analyze the deviant text until it aligns with our specified processing standards.

\setlength{\abovecaptionskip}{0.1cm}
\setlength{\belowcaptionskip}{-0.3cm}
To identify the performance-critical knobs for subsequent tasks, we prefer that knobs mentioned more frequently in the extracted tuning hints are usually more critical. Therefore, we count the mentioned times for each knob and then accordingly rank and select the performance-critical ones. The value ranges of these performance-critical knobs are determined jointly by the official configuration specification and our deployment environment to ensure the search space is safe and valid.

\section{Hint-Aware Demonstration Reinforcement Learning}\label{sec-5}
\subsection{DRL Formulation for Knobs Tuning}\label{sec:5-1}
First, we translate the knobs tuning problem into a deep reinforcement learning (DRL)
formulation. Considering the limit of space, here we only introduce the essential components:\
\begin{itemize}[leftmargin=*]
    \item $\textbf{State:}$ State $s_t$ captures the system's current status. Specifically, it is a vector composed of runtime metrics from both the operating system and the target DBMS such as CPU utilization, cache hit blocks, etc.

    \item \textbf{Action:} Action $a_t$ is a vector where each dimension represents a performance-critical knob of the target DBMS, with values normalized to [-1,1]. Here, -1 denotes the minimum setting, 1 the maximum, with intermediate values proportionally scaled according to the value range.

    \item $\textbf{Reward:}$ Reward $r_t$ quantifies the effectiveness of the generated action $a_t$ under current state $s_t$. It reflects the performance improvement achieved by current knobs configuration compared to the default configuration.

    \item \textbf{Policy:} Policy \( \pi(s_{t} \lvert \theta^{\pi}) \) maps a given state to an action. It is implemented in DRL as a deep neural network that processes the  state as input and outputs  knobs configuration. For each tuning iteration, the DRL agent is responsible for generating a promising action based on the current state and the latest policy.

    \item \textbf{Transition:} A transition records the state change resulting from an action, denoted as \( (s_t, a_t, r_t, s_{t+1}) \). Transitions are used to update value functions and policy networks in DRL.\
\end{itemize}



\subsection{Discussions About Demonstration Reinforcement Learning}
During the training process of DRL, the policy network learns to select actions by optimizing its parameters based on  environmental feedback, while the value function estimates the expected rewards for these actions, guiding the policy network through transitions to maximize cumulative rewards. Without domain knowledge, the training process usually has to start from scratch and thus suffers from the cold start problem. Unfortunately, conventional DRL algorithms including Deep Q-Networks (DQN)~\cite{mnih2015human}, Policy Gradients~\cite{schulman2015high} and Actor-Critic~\cite{lillicrap2015continuous} cannot directly integrate domain knowledge into the training phase. To address this issue, demonstration reinforcement learning algorithms such as DDPGfD (Deep Deterministic Policy Gradient from Demonstrations)  have been recently developed~\cite{vecerik2017leveraging}. These algorithms can integrate demonstration experiences into the learning process, effectively bridging the gap between pure reinforcement learning and imitation learning.

As far as we know, DemoTuner is  the first to introduce demonstration reinforcement learning to achieve efficient DBMS knobs tuning. However, it is worth noting that some tuning hints' utilities tend to diminish with training steps, in the meanwhile, transitions with real environmental feedback may provide more information than the static domain knowledge as training progresses. Consequently, treating demonstrations from domain knowledge on par with other transitions, as conventional demonstration reinforcement learning algorithms did, may miss opportunities to further improve DBMS's performance. In order to effectively integrate the tuning hints into the entire DRL training phase, we propose a hint-aware demonstration reinforcement learning method named HA-DDPGfD in DemoTuner. Specifically, HA-DDPGfD initially utilizes the demonstrations derived from tuning hints to pre-train the agent (Section~\ref{sec:5-3}). To appropriately guide the subsequent fine-tuning phase with the mined tuning hints, we introduce a hint priority-driven prioritized experience replay mechanism as well as a hint priority update policy (Section~\ref{sec:5-4}) in HA-DDPGfD. Moreover, we also develop a hint-guided reward shaping technique (Section~\ref{sec:5-5}) to further accelerate the convergence speed of agent training.

\subsection{Pre-Training Agent with Demonstrations}\label{sec:5-3}
\subsubsection{Initial Matched Hints Selection} Initially, we should filter out the tuning hints matched with the target DBMS's current (default) runtime environment. To accomplish this, we leverage Prometheus and DBMS's monitoring capabilities to collect runtime metrics under the default knobs configuration. Only those tuning hints whose $condition\_vector$ coincide with the DBMS's default runtime environment are selected to form the initial $Match\_Hints$ set. Since the benefits of tuning hints can vary throughout the training processes, we assign a dynamic priority $prior_i$ to each hint $h_i$ in $Match\_Hints$ to  indicate its relevance in the current training iteration. 



\subsubsection{Demonstrations Collection} In order to derive demonstrations from the matched tuning hints, we first use the \emph{rec\_action} in each hint $h_i$ to modify the default value of the corresponding knob $knob_{h_i}$ while leaving other knobs as default. This one-hint-participation manner can prevent multiple hints from simultaneously influencing the knob settings, allowing us evaluate each hint's contribution to performance improvement. Specifically, the knob value is modified as follows:
\setlength{\abovedisplayskip}{0.1cm}
\begin{equation}
\label{eq:1}
knob_{h_i} =
\begin{cases}
value_{h_i}, & \text{if $h_i$ is explicit} \\
knob_{h_i} \pm k \cdot \left( e^{(n+1) \cdot C} - e^{n \cdot C} \right), & \text{if $h_i$ is implicit}
\end{cases}
\end{equation}
If $rec\_action$ is an explicit recommended value $value_{h_i}$ in $h_i$, e.g., 1GB or 25\% of memory, we can calculate the corresponding value and $knob_{h_i}$ is updated to $value_{h_i}$. On the other hand, if $rec\_action$ is an implicit expression such as knob increase or decrease, we should modify the value according to the tuning direction as well as current value of $knob_{h_i}$. Instead of applying a fixed adjustment change step, we employ a logarithmic scale to partition the value range, avoiding overly aggressive changes under broad value ranges while maintaining
fine-grained control under narrower ranges. Specifically, the exponential scaling factor $C$ is computed as $C = \ln(\max - \min)/z$ which divides the value range into $z$ logarithmic intervals. $n$ is the interval index in which the current value of $knob_{h_i}$ falls and $k$ is a coefficient determines the adjustment magnitude. 


With the adjusted knobs configuration $kc_{adj}$ according to tuning hint $h_i$, we evaluate the performance and collect runtime metrics to obtain a demonstration. In detail, we use a quintuple $<$$s_{def}, kc_{adj}, perf_{imp}, s'_{def}, h_i$$>$ to denote a demonstration, which means that after the default knobs configuration is adjusted according to tuning $h_i$, the state changes to $s'_{def}$ and DBMS achieves a $perf_{imp}$ performance improvement ratio over the default performance $perf_0$. That is to say, a demonstration consists of a transition part and a corresponding tuning hint part. These achieved demonstrations will be stored into the experience replay buffer before training begins and retained throughout the entire training process. Finally, the priority $prior_i$ of tuning hint $h_i$ will also be updated to the performance improvement ratio $perf_{imp}$ to measure the current benefit of $h_i$.





\subsubsection{Agent Pre-Training} To mitigate the ``search from scratch'' cold-start problem encountered by common DRL agent training tasks, we choose to leverage the demonstrations generated from the extracted tuning hints to pre-train the DDPG agent.
Specifically, the actor and critic networks are initialized via gradient updates using mini-batches sampled from the experience replay buffer, which contains only demonstrations at this stage.
With the pre-trained networks, the agent is actually guided by the tuning hints and thus gains a certain capability to recommend good knob configurations.

\subsection{Fine-Tuning Agent with Hint Priority-Aware PER}\label{sec:5-4}
Based on the warm-start capability of the pre-trained agent, we can generate actions and obtain transitions based on actual configuration evaluation. To achieve an adaptive utilization of demonstrations and transitions as training
progresses, we propose a hint priority-aware prioritized experience replay (PER) mechanism to fine-tune the agent. We also propose a hint priority update policy to dynamically adjust hints' priorities along with the fine-tuning process.

\subsubsection{Hint Priority-Aware PER}
Prioritized experience replay~\cite{schaul2015per} is a key technique to achieve demonstration reinforcement learning. It assigns a priority to each experience (including both demonstrations and transitions) based on its significance, which affects its probability of being sampled for model updates. In detail, the probability of sampling a particular experience $i$ from the experience replay buffer is defined as:
\begin{equation}
\label{deqn_ex1}
P(i)=\frac{p_{i}^{\alpha } }{\sum_{m} p_{m}^{\alpha }} ,
\end{equation}
where $p_{i}^{\alpha }$ is the priority of experience $i$ raised to the power of $\alpha$ and $\alpha$ is a parameter that determines how much prioritization is used. In PER, the priority $p_{i}$ of each  experience is primarily determined by its Temporal Difference (TD) error, which reflects the difference between the predicted and the actual outcomes. However, since the benefits of demonstrations may diminish over training iterations, we propose a hint priority-aware PER mechanism named hpPER.\ Specifically, the priority of each  experience in hpPER can be calculated as follow:

\setlength{\abovedisplayskip}{0.0cm}
\setlength{\belowdisplayskip}{0.2cm}
\begin{equation} \label{eq-3}
p_i = \delta^{2} + \lambda_{1} |\nabla_{\alpha} Q(s_i, a_i)|\theta^Q|^{2} + \epsilon + \lambda_{2}prior_{j}
\end{equation}

This enhanced priority formulation goes beyond traditional reinforcement learning approaches that focus solely on TD error. In Eq.\ (\ref{eq-3}), $\delta^2$ represents the TD error. The second term quantifies the loss function of actor, with its magnitude signaling the experience's significance for policy network updates. The third term $\epsilon$ is a small positive constant ensuring that all experiences have a non-zero probability of being selected,  thus promoting diversity in the training data. The novel inclusion, $prior_{j}$ in the last term indicates the dynamic priority of the tuning hint $h_j$ related to current demonstration $i$, which enhances the sampling probability of demonstrations according to the latest priorities of their linked specific tuning hints. Besides, we use $\lambda_{2}$ to control the enhancement ratio. Detailed discussion about the value of  $\lambda_{2}$ can be found in Section~\ref{sec:7-d-2}. For transitions in the experience replay buffer, the last term is 0. By incorporating these diverse factors, hpPER is able to assign appropriate priorities to each transition and demonstration for current training step, thus effectively integrating the domain knowledge into the DRL training phase.

\subsubsection{Hint Priority Update}
The hint priority $prior_j$ is used to measure current benefit in performance improvement of tuning hint $h_j$ at training step $t$. As outlined in Eq.\ (\ref{eq-3}), this metric is crucial for assessing the importance of demonstrations in the experience replay buffer. However, as training progresses, some tuning hints may no longer have the same good effect as the beginning. Therefore, DemoTuner should be able to dynamically update the hint priority throughout the training process. To this end, we first leverage an active action adjustment technique to modify the generated action $a_t$ by the actor's policy network $\pi(s_{t-1}, \lvert \theta^{\pi})$. Specifically, we first obtain current $Match\_Hints$ set according to the application condition $condition\_vector$ of tuning hints as well as DBMS's current runtime environment. After that, we probabilistically select one tuning hint $h_j$ according to hints' current priorities from $Match\_Hints$ and adjust the corresponding knob value according to Eq.\ (\ref{eq:1}) while leaving the others unchanged. On the one hand, this modification facilitates the direct involvement of domain knowledge during agent training. On the other hand, it utilizes the real interaction with the target environment and thus can help accurately measure the priority of the selected hint $h_j$. According to the environmental feedback to the adjusted action, we then update $h_j$'s priority as follows:
\setlength{\abovedisplayskip}{0.1cm}
\setlength{\belowdisplayskip}{0.1cm}
\begin{equation}
    prior'_{j} = \frac{1}{2} \left( prior_j + \frac{perf_t - perf_0}{perf_t} \right),
\end{equation}
where \(perf_t \) denotes the performance of the adjusted action, and \( perf_0 \) represents the performance under default configuration. This dynamic hint priority update policy ensures that valuable demonstrations in the experience replay buffer are adequately emphasized, while outdated or misleading demonstrations gradually diminish in influence along with the agent fine-tuning process.

\subsection{Hint-Guided Reward Shaping}\label{sec:5-5}
Reward is one of the most crucial components in RL.\ Reward shaping~\cite{ng1999rewardshaping} involves providing small intermediate synthetic rewards to the agent to expedite training convergence. In HA-DDPGfD, we design a hint-guided reward shaping technique to leverage the domain knowledge to accelerate the agent fine-tuning phase. Specifically, after obtaining the original reward $r_t$ defined as the performance improvement over default $perf_0$ (as described in Eq.\ (\ref{eq:reward})), we adjust $r_t$ according to the divergence between the actual knob adjustments and the actions $rec\_action$ suggested by the tuning hints in current $Match\_Hints$ set. For instance, if a tuning hint suggests increasing the value of a specific knob but it actually decreases from action $a_{t-1}$ to action $a_t$, this discrepancy is treated as a violation of the tuning hint. It is worth noting that we only take the current matched tuning hints with priorities larger than a specified threshold into consideration. In detail, we use $\sum\mathbb{I}(a_t, a_{t-1}, h_i)$ (where $h_i\in Match\_Hints$) to count the total number of violations, with $\mathbb{I}(\cdot)$ being the indicator function that takes the value of 1 when a violation occurs, and 0 otherwise. Then the shaping factor $f_t$ is obtained as:
\begin{equation}
\label{equ:reward-shaping}
f_t = \beta \cdot \frac{\sum_{h_i \in Match\_Hints} \mathbb{I}(a_t, a_{t-1}, h_i)}{\vert Match\_Hints  \vert},
\end{equation}
where $\vert Match\_Hints  \vert$ is the number of current matched tuning hints. Besides, \( \beta \) is a coefficient to control the extent of reward shaping. The magnitude of this coefficient significantly impacts the effectiveness of agent training. If it is too large, it may destabilize the training process, while if it is too small, it may fail to sufficiently accelerate the training. A detailed discussion regarding the setting of  \( \beta \) can be found in Section~\ref{subsubsec:reward_shaping_ratio q}. Finally, the shaped reward $r'_t$ can be calculated as follows:
\setlength{\abovedisplayskip}{0.1cm}
\begin{subequations}\label{eq:combined}
    \begin{align}
        r_t &= \frac{perf_{t} - perf_{0}}{perf_{0}}, \label{eq:reward} \\
        r'_t &= \left(1 - f_t\right) \cdot r_t \label{eq:rewards}
    \end{align}
\end{subequations}

This hint-guided reward shaping technique is able to strike a balance between domain knowledge utilization and autonomous exploration by the agent, thereby optimizing training performance through efficiently integrating tuning hints.

\subsection{The Proposed HA-DDPGfD Algorithm in DemoTuner}
The detailed workflow of HA-DDPGfD in DemoTuner is outlined in Algorithm~\ref{HA-DDPGfD}. In the agent pre-training phase (Lines 3--13), DemoTuner collects demonstrations through adjusting the default configuration based on matched tuning hints and conducting configuration evaluation. With these demonstrations stored in the experience replay buffer, DemoTuner pre-trains the agent's critic network according to the 1-step loss: $L^{1}_{Critic}(\theta^Q) = \frac{1}{2}{(r_t - Q(s, \pi(s)|\theta^Q))}^2$, and updates the policy network according to the following
policy gradient equation:
\begin{equation}
\nabla_{\theta^\pi} L_{Actor}(\theta^\pi) = -  \nabla_{\theta^\pi} J(\theta^\pi) + \beta_{2} \nabla_{\theta^\pi} L_{reg}^A(\theta^Q)
\end{equation}

\begin{algorithm}[t]
\small
\caption{HA-DDPGfD in DemoTuner}\label{alg:alg1}
\begin{algorithmic}[1]\label{HA-DDPGfD}
\STATE \textbf{Input:} Mined tuning hints \( H \), Priority replay buffer \( P \), DBMS's runtime environment \( Env \)
\STATE \textbf{Output:} Action-value function \( Q(\cdot \mid \theta^Q) \) (critic) and policy \( \pi(\cdot \mid \theta^\pi) \) (actor) for online knobs tuning

\centerline{\textbf{\noindent \rule[2pt]{1cm}{0.7pt} Offline Training Phase 1: Pre-Training \noindent \rule[2pt]{1cm}{0.7pt}}}

\STATE Initialize state $s_{def}$ with default $Env$
\STATE Obtain current $Match\_Hints$ from $H$ based on $s_{def}$
\WHILE{unused tuning hints in $Match\_Hints$ $\neq \emptyset$}
        \STATE Randomly sample $h_i$ from unused tuning hints in $Match\_Hints$
        \STATE Generate knobs configuration $kc_{adj}$ via adjusting the value of knob $knob_{h_i}$ according to Eq. (\ref{eq:1})
        \STATE Evaluate the performance of $kc_{adj}$ and obtain state $s'_{def}$, performance improvement $perf_{imp}$
        \STATE Store the demonstration $(s_{def}, kc_{adj}, perf_{imp},  s'_{def}, h_{i})$ to experience buffer $P$
        \STATE Update $prior_i$ for $h_i$ according to $perf_{imp}$
        \STATE mark $h_i$ as used
\ENDWHILE
\STATE Pre-train agent's networks with $L^{1}_{\text{Critic}}(\theta^{Q})$ and $L_{\text{Actor}}(\theta^{\pi})$ based on buffer $P$

\centerline{\textbf{\noindent \rule[2pt]{1cm}{0.7pt} Offline Training Phase 2: Fine-Tuning \noindent \rule[2pt]{1cm}{0.7pt}}}

\FOR{each episode $e = 0$ to $M$}
    \FOR{each step $t = 0$ to $T$}
        \STATE Update $Match\_Hints$ based on $s_{t}$
        \STATE Generate a new action $a_{t} = \pi (s_{t}, \theta^{Q}) + n_{t}$
        \STATE Probabilistically select a tuning hint $h_i$ from $Match\_Hints$ based on priorities
        \STATE Adjust $a_{t}$ via modifying the value of knob $knob_{h_i}$ according to Eq. (\ref{eq:1})
        \STATE Evaluate the performance of $a_t$ and obtain state $s_{t+1}$, reward $r_t$ according to Eq. (\ref{eq:reward})
         \STATE Update $prior_i$ for $h_i$
       \STATE Shape $r_t$ according to Eq. (\ref{eq:rewards}) and obtain $r'_t$
        \STATE Store both 1-step and \( n \)-step transitions in buffer \( P \)
    \ENDFOR
    \STATE Sample a minibatch of transitions with the hpPER mechanism, calculate $L^{1}_{\text{Critic}}(\theta^{Q})$ and $L^{n}_{\text{Critic}}(\theta^{Q})$
        \STATE Update the agent with $L_{\text{Critic}}(\theta^{Q})$ and $L_{\text{Actor}}(\theta^{\pi})$
\ENDFOR
\end{algorithmic}
\end{algorithm}

Then during the agent fine-tuning phase (Lines 14--27), DemoTuner iteratively interacts with the target DBMS via generating promising configurations and obtaining performance feedback to improve the policy. For each step, DemoTuner generates an action according to current policy network and then adjust it based on the hints' priorities. After that, the adjusted action is evaluated and both 1-step and n-step transitions are stored in the experience replay buffer.
For each episode, DemoTuner follows the proposed hpPER mechanism to select experiences to update the agent's critic and actor networks. Specially, to better balance the short-term and long-term reward feedback, we also leverage a mix of 1-step and n-step returns to update the critic network as previous work~\cite{vecerik2017leveraging} did. Specifically, the n-step return is defined as follows:
\begin{equation}
R_{n} = \sum_{i=0}^{n-1}\gamma^{i} r_{i}+\gamma^{n}Q(s_{n-1}^{'},\pi(s_{n-1}^{'};\theta ^{Q^{'}} ) ),
\end{equation}
where \( r_i \) is the immediate reward at the \( i \)-th step, \( \gamma \) is the discount factor, and \( Q(s_{n-1}^{'}, \pi(s_{n-1}^{'}, \theta^{Q^{'}})) \) is the next-step $Q$-value calculated by the critic network based on the current policy and state. Then the corresponding n-step loss function is defined as:
\begin{equation}
L_n(\theta^Q) = \frac{1}{2}{(R_n - Q(s, \pi(s)|\theta^Q))}^2
\end{equation}
This loss function minimizes the gap between the $Q$-value and the n-step return, optimizing the critic network's $Q$-value estimation. Finally, a combined loss function $L_{critic}(\theta^Q)$ that integrates both 1-step and n-step losses is defined as:
\begin{equation}
L_{critic}(\theta^Q) = L_1(\theta^Q) + \beta_1 L_n(\theta^Q) + \beta_2 L_{reg}^C(\theta^Q),
\end{equation}
where \( L_1 \) and \( L_n \) are respectively the 1-step and n-step losses, and \( L_{reg}^C \) is the regularization term for the critic network, while \( \beta_1 \) and \( \beta_2 \) are weight hyperparameters.

\section{Experimental Setups}\label{sec-6}
\subsection{Experimental Platform}
To evaluate the effectiveness and efficiency of DemoTuner, we conducted a series of experiments using two widely utilized database management systems MySQL (version 8.0.36) and PostgreSQL (version 16.1). For configuration evaluation, we deploy the target DBMS on a  Linux server with an Intel processor of 8 physical cores, 32GB of RAM, and a 300GB SSD.\ The operating system is Ubuntu 22.04.2 with kernel version 6.5.0. In order to mitigate the potential performance interference, we make sure that nothing else is running on the server except the essential kernel processes and the target DBMS service when conducting performance tests to evaluate generated configurations. It is worth noting that DemoTuner currently supports two widely deployed DBMSs: MySQL and PostgreSQL.\ However, we can easily add supports for other DBMSs according to the following two steps. First, for tuning hints extraction, we need to collect source texts about knobs tuning for the new DBMS and change the target DBMS information in the prompt. Second, for offline agent training, we only need to modify the control script for automatic configuration evaluation of the new DBMS.\ The source code of DemoTuner is publicly available at~\cite{DemoTuner2025Codes} for further extension and improvement.



\subsection{Benchmark Applications}

The main benchmark we employed for configuration evaluation is YCSB (Yahoo! Cloud Serving Benchmark) \cite{cooper2010benchmarking}, which is a frequently-used benchmarking tool for performance tests of DBMSs. Similar to WATuner \cite{ge2021watuning}, we also synthesize different workloads based on the proportions of operations (read, update, etc.) in YCSB. Specifically, we evaluate DemoTuner as well as the baseline methods with three different workload types including Read-Heavy (RH), Read-Write (RW), and Write-Heavy (WH), and list the concrete operation proportions in Table \ref{YCSB_Workload_Type}. For each workload type, we generate 1,000,000 records for PostgreSQL and 500,000 records for MySQL, and each configuration evaluation involves 500,000 operations launched with 10 concurrency threads. The workload datasize and operation number are set according to the computing capability of our testbed as well as considering the time consumption of different experiments.


\begin{table}[t]
    \centering
    \caption{The details
 of synthetic workloads with YCSB.}

    \renewcommand{\arraystretch}{1.5} 
    \setlength{\tabcolsep}{6pt} 

    \resizebox{0.9\linewidth}{!}{  
    \begin{tabular}{c|c|c|c|c}
    \Xhline{1.2pt}
    \textbf{Workload} & \textbf{READ} & \textbf{UPDATE} & \textbf{INSERT} & \textbf{SCAN} \\
    \hline
    \text{Read-Heavy (RH)} & 0.6 & 0.1 & 0.1 & 0.2 \\
    \hline
    \text{Read-Write (RW)} & 0.4 & 0.2 & 0.3 & 0.1 \\
    \hline
    \text{Write-Heavy (WH)} & 0.1 & 0.4 & 0.4 & 0.1 \\
    \Xhline{1.2pt}
    \end{tabular}
    }\label{YCSB_Workload_Type}
\end{table}



In addition to YCSB, TPC-H (OLAP) and TPC-C (OLTP) workloads are also used to evaluate the adaptability of DemoTuner to unknown workloads in Section~\ref{adaptive of different workload} and provide a query-level performance improvement analysis in Section~\ref{subsec:query-level}. According to GPTuner~\cite{lao2025gptuner}, we set the scaling factor of TPC-H and TPC-C to be 1 and 200, respectively. Besides, the TPC-C benchmark operates with 32 terminals and a transaction rate of 50,000. These workload settings are implemented with BenchBase~\cite{difallah2013oltpbench}.


\subsection{Selected Knobs and Tuning Hints}
In our experiments, we use a set of search keywords such as \emph{PostgreSQL performance tuning} and \emph{MySQL configuration tuning} to collect source texts from various documents such as web forums, blog posts, official manuals and configuration files. GPT-3.5-Turbo~\cite{gpt2025} is utilized considering the tradeoff between costs and  practical effectiveness. After tuning hints extraction, we prioritize the knobs according to their mentioned frequency since this metric well indicates the degree of correlation with performance. We also consult official configuration guides as well as empirical data to further adjust our selection. Following the above process, we finally select 19 performance-critical knobs for MySQL and 18 for PostgreSQL.\ Besides, we obtain 70 tuning hints (average 3.7 hints per knob) for MySQL and 104 (average 5.8 hints per knob) for PostgreSQL.\

\subsection{Baseline Methods}
To evaluate the performance of DemoTuner in DBMS knobs tuning, we select two recent related studies as the baselines which also tried to combine domain knowledge with machine learning methods, namely DB-BERT~\cite{trummer2023db} and GPTuner~\cite{lao2025gptuner}. Additionally, we also compare DemoTuner with CDBTune~\cite{zhang2019CDBTune}, a representative method that uses deep reinforcement learning for knobs configuration. In our experiments, DemoTuner undergoes a total 200 iterations of configuration evaluation for agent pre-training and fine-tuning, with another 10 iterations conducted for online tuning. As follows, we give a brief introduction of each baseline and how we implement them.
    \begin{itemize}
        \item DB-BERT~\cite{trummer2023db} is a recent NLP-enhanced online knobs tuning method which utilizes BERT to translate natural language hints into recommended settings. At runtime, it employs an iterative RL algorithm to aggregate, adapt, and prioritize these hints for guiding the selection of knobs settings to evaluate. We utilize the available implementation version from the authors and set the RL iteration number to be 210, which is equal to the total number of offline and online iterations of DemoTuner.

        \item GPTuner~\cite{lao2025gptuner} first utilizes a LLM-based pipeline to collect and refine domain knowledge to achieve efficient knob selection and value range identification. It then explores the optimal knob settings using a coarse-to-fine Bayesian optimization framework, which sequentially explores two search spaces of different granularity. We use the authors' version and leverage 200 initial sampling configurations (30 in the coarse manner and 170 in the fine manner) recommended by GPTuner to construct the initial Gaussian process model. The online tuning iteration number is also set to be 10.

         \item CDBTune~\cite{zhang2019CDBTune} is an end-to-end automatic cloud database tuning system that employs deep deterministic policy gradient methods to find the optimal configuration in high-dimensional continuous spaces. We use the PyTorch library to build neural networks to implement DDPG in CDBTune. It also undergoes 200 iterations for agent offline training, with another 10 iterations conducted for online tuning.

    \end{itemize}



\section{Experimental Results and Analysis}\label{sec-7}



\begin{table*}[t]
    \centering
    \caption{Comparison of online tuning cost (seconds) consumed by DemoTuner and the baselines.}\label{tab:cost_comparison}
    \renewcommand{\arraystretch}{1.3}
    \setlength{\tabcolsep}{10pt}

    \begin{tabular}{c|c|c|c|c|c}
        \Xhline{1.2pt}
        \textbf{Target DBMS} & \textbf{Workload Type} & \textbf{DemoTuner} & \textbf{CDBTune} & \textbf{DB-BERT} & \textbf{GPTuner} \\
        \hline
        \multirow{3}{*}{MySQL} & RH & \textbf{627.07 } & 752.13  & 699.27  & 825.38  \\
        & RW & \textbf{599.96 } & 660.66  & 743.32  & 1126.37  \\
        & WH & \textbf{806.01 } & 897.56  & 927.02  & 896.86  \\
        \hline
        \multirow{3}{*}{PostgreSQL} & RH & \textbf{401.00 } & 420.51  & 593.97  & 457.65  \\
        & RW & \textbf{311.34 } & 399.50  & 463.70  & 327.10  \\
        & WH & \textbf{341.73} & 351.26 & 444.10 & 532.99  \\
        \Xhline{1.2pt}
    \end{tabular}
\end{table*}

\subsection{Performance Comparison}
\label{sec:6-1}
\subsubsection{Effectiveness}



\begin{figure}[t]
    \centering
      \includegraphics[width=1\linewidth]{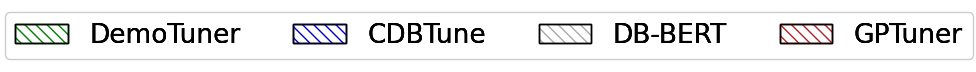}
     \begin{minipage}{0.32\linewidth}
        \centering
        \includegraphics[width=1\textwidth]{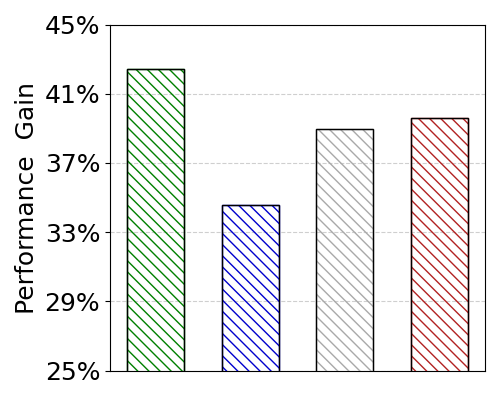}
        \subcaption{MySQL RH}\label{fig:mysql_RW} 
    \end{minipage}
    \hfill
    \begin{minipage}{0.32\linewidth}
        \centering
        \includegraphics[width=1\textwidth]{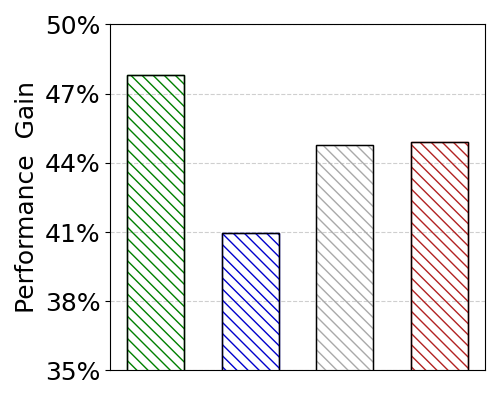}
        \subcaption{MySQL RW}\label{fig:postgresql_WO} 
    \end{minipage}
    \hfill
    \begin{minipage}{0.32\linewidth}
        \centering
        \includegraphics[width=1\textwidth]{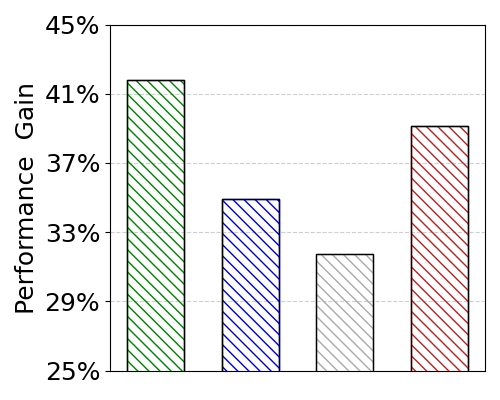}
        \subcaption{MySQL WH}\label{fig:mysql_WH} 
    \end{minipage}

    \vspace{0.5cm} 

    \begin{minipage}{0.32\linewidth}
        \centering
        \includegraphics[width=1\textwidth]{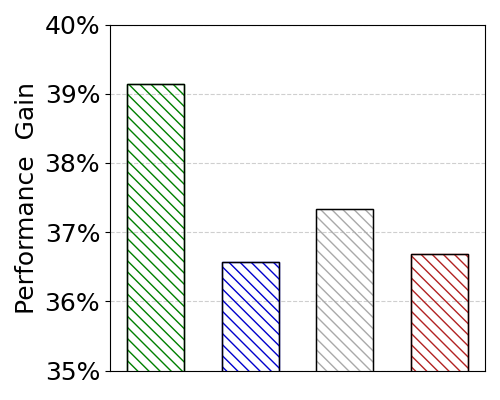}
        \subcaption{PostgreSQL RH}\label{fig:postgresql_RH} 
    \end{minipage}
    \hfill
    \begin{minipage}{0.32\linewidth}
        \centering
        \includegraphics[width=1\textwidth]{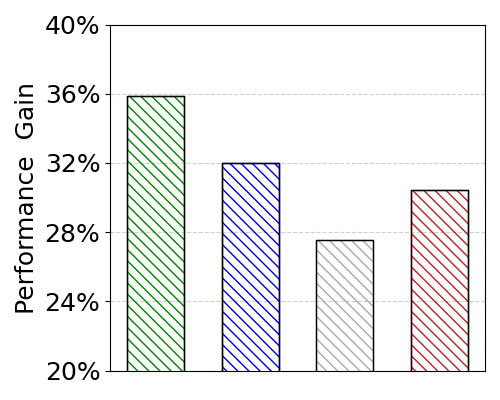}
        \subcaption{PostgreSQL RW}\label{fig:mysql_RH} 
    \end{minipage}
    \hfill
    \begin{minipage}{0.32\linewidth}
        \centering
        \includegraphics[width=1\textwidth]{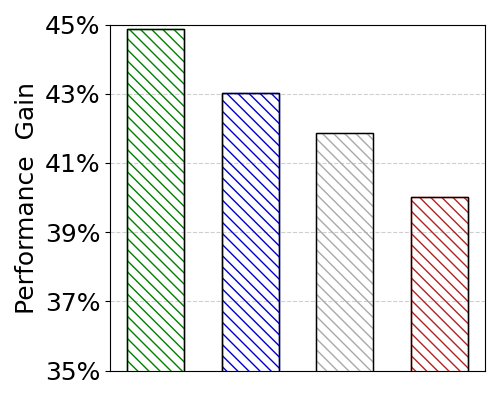}
        \subcaption{PostgreSQL WH}\label{fig:postgresql_RW} 
    \end{minipage}

    \caption{The best performance gains achieved by DemoTuner and the baselines across different workload types on PostgreSQL and MySQL, respectively.}\label{fig:performance_comparison}
\end{figure}

In order to evaluate the effectiveness of DemoTuner for knobs tuning, we utilize DemoTuner as well as the three baselines to tuning knobs of PostgreSQL and MySQL under the three different workload types listed in Table~\ref{YCSB_Workload_Type}. The performance gains over the default performance $perf_0$ achieved by the best configurations found by different methods are shown in Figure~\ref{fig:performance_comparison}. Specifically, the best performance gain is calculated as:

\begin{equation}
    \nonumber
    PG_{\text{best}} = \frac{perf_{0} - perf_{\text{best}}}{perf_{0}} \times 100\%.
\end{equation}

We can observe that DemoTuner can always find the best configuration with the largest performance gain over default across different workload types and DBMSs. The average best performance gain over default across three different workloads achieved by DemoTuner reaches as much as 44.01\% for MySQL and 39.95\% for PostgreSQL, respectively.
Since CDBTune does not utilize any tuning hints to accelerate the initial training stage of the DDPG algorithm, it fails to train a comparable agent with DemoTuner when the experience replay buffer consists of a certain number of low-performance transitions. As a result, DemoTuner is able to further reduce the default execution time by 6.86\% to 7.85\% for MySQL workloads while 1.86\% to 3.83\% for PostgreSQL workloads compared with CDBTune. The performance of DB-BERT heavily relies on the quality of initial tuning hints extraction, however, totally neglecting the application condition of tuning hints will mislead the offline training and online tuning tasks. Besides, DB-BERT is only able to explore a limited range around the recommended knob values by the tuning hints, which may miss opportunities to further improve the performance. Therefore, compared with DB-BERT, DemoTuner can further reduce the default execution time by 10.03\% at most under the MySQL Write-Heavy workload and 8.31\% at most under the PostgreSQL Read-Write workload with the same constraint of online tuning iterations. Although GPTuner enhances the selection of critical knobs and the generation of value ranges with tuning hints, it cannot ensure the accurate alignment between domain knowledge and current runtime environment since Bayesian optimization inherently lacks the ability to incorporate system states into its iterative process. In contrast, DemoTuner utilizes demonstrations to accelerate the pre-training stage and leverages the HA-DDPGfD algorithm to effectively integrate the extracted tuning hints throughout the agent’s fine-tuning process. In our experiments, the average best performance gain over default achieved by GPTuner is  41.21\% for MySQL workloads and 35.71\% for PostgreSQL workloads. That is to say, DemoTuner can further reduce the default execution time by respectively an average 2.80\% for MySQL and 4.24\% for PostgreSQL compared with GPTuner.

\setlength{\abovecaptionskip}{0.4 cm}
\setlength{\belowcaptionskip}{-0.3 cm}

\subsubsection{Online Tuning Cost}
In addition to the achieved performance gains, the total tuning costs consumed during the online knobs tuning phase should also be taken into consideration. Here we define the online tuning cost as the total execution time of the 10 online tuning iterations of each approach. For DB-BERT, the last 10 iterations are considered. Notably, we exclude the time consumed by model updates and configuration deployment from our analysis, as their impact is significantly smaller than that of execution time. As indicated in Table~\ref{tab:cost_comparison}, DemoTuner not only delivers the highest performance improvements but also incurs the lowest online tuning costs across various workload types for both PostgreSQL and MySQL. It is worth noting that the average online tuning cost consumed by the pure DDPG-based method CDBTune is 29.54\%, 11.66\% less than that of DB-BERT and GPTuner on PostgreSQL and 3.63\%, 22.44\% on MySQL, respectively. Unlike DB-BERT and GPTuner, CDBTune can explore the entire configuration space and is responsive to the dynamic system runtime environment, thus avoiding the pitfalls of misleading tuning hints.  Compared with CDBTune, DemoTuner employs  demonstration reinforcement learning to incorporate domain knowledge into the agent's offline training phase, thereby further enhancing online tuning performance. Consequently, DemoTuner consumes an average 17.55\% less online tuning cost on MySQL and an average 14.37\% less online tuning cost on PostgreSQL than CDBTune.

\subsection{Adaptability of DemoTuner to Unknown Workloads}\label{adaptive of different workload}
In practical scenarios, the workload characteristics faced by a database management system typically change over time, which makes training separate offline agents for each workload type costly. Hence, DemoTuner may have to tuning knobs for previously unknown workloads. To evaluate the adaptability of DemoTuner  and the baselines, we use the agents respectively trained under TPC-H and the YCSB Write-Heavy workload to recommend configurations for the unknown TPC-C workload. All experiments are conducted with PostgreSQL and we use the 99th-tile latency as the performance metric.

\begin{figure}[t]
    \centering
     \includegraphics[width=0.49\textwidth]{pictures/Exper1/legend.png}
	\begin{minipage}{0.49\linewidth}
		\centering
		\includegraphics[width=1\textwidth]{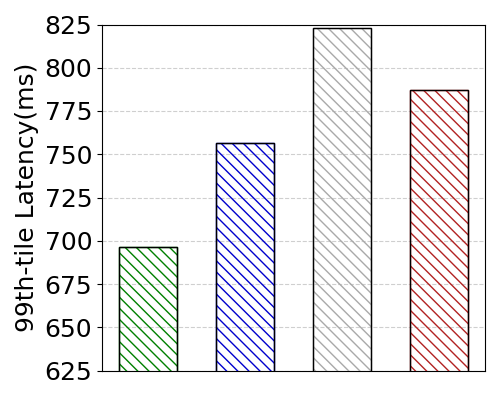}
		\subcaption{TPC-H$\rightarrow$TPC-C}\label{fig:TPCH->TPCC} 
	\end{minipage}
	\begin{minipage}{0.49\linewidth}
		\centering
		\includegraphics[width=1\textwidth]{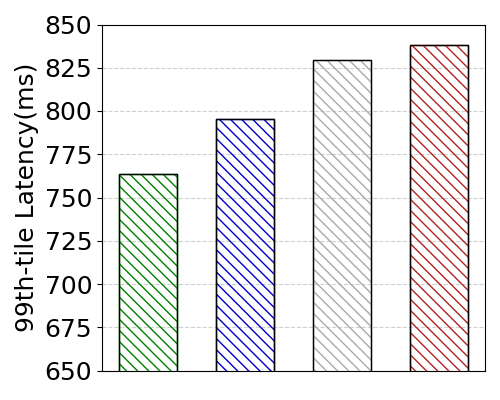}
		\subcaption{YCSB (WH)$\rightarrow$TPC-C}\label{fig:YCSB->TPCC} 
	\end{minipage}
    \caption{Comparison of adaptability between DemoTuner and the baselines. }\label{fig:adaptability}
\end{figure}

As shown in Figure~\ref{fig:adaptability}, with the same 10 online tuning iterations, DemoTuner can still always find the best configuration among both application scenarios. Specifically, under the TPC-H $\rightarrow$ TPC-C scenario, DemoTuner can reduce the 99th-tile latency by 7.92\%, 15.38\% and 11.54\% compared to CDBTune, DB-BERT and GPTuner, respectively. In addition, under the YCSB (WH) $\rightarrow$ TPC-C scenario with more significant difference, DemoTuner can still reduce the latency by 4.03\%, 7.99\% and 8.90\% compared to CDBTune, DB-BERT and GPTuner, respectively. In detail, DB-BERT does not consider the application condition of tuning hints at all and solely relies on the accuracy of extracted tuning hints, as a result, it is difficult to achieve a satisfying performance under an unknown workload. Although GPTuner utilizes a workload-aware manner to identify critical knobs and their value ranges, however, since Bayesian optimization inherently lacks the ability to capture changes in environments, GPTuner has to directly use the surrogate model trained on the source workloads to execute the online knobs tuning task under the unknown target workload. In contrast, with the RL agent's ability to perceive the changes in runtime environments and generate actions accordingly, CDBTune does not need to establish a new model and owns
a good adaptability when the workload changes are not very large. Compared with CDBTune, DemoTuner is able to leverage the tuning hints according to their application conditions and adaptively adjust the hints' priorities to align the direction of action directions with the current runtime environment, and thus achieving a further performance improvement under both scenarios.

\setlength{\abovecaptionskip}{0.4 cm}
\setlength{\belowcaptionskip}{-0.4 cm}




\subsection{Ablation Experiments}\label{sec:6-2}





\subsubsection{Effectiveness of Hint Priority Update} 
\begin{figure}[t]
\includegraphics[width=0.6\linewidth]{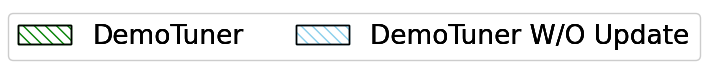}
    \centering
	\begin{minipage}{0.49\linewidth}
		\centering
		\includegraphics[width=1\textwidth]{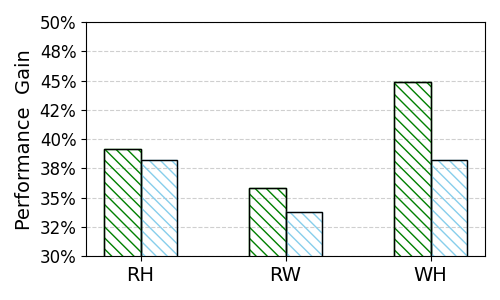}
		\subcaption{Best performance gain}\label{fig:action guide performace} 
	\end{minipage}
	\begin{minipage}{0.49\linewidth}
		\centering
		\includegraphics[width=1\textwidth]{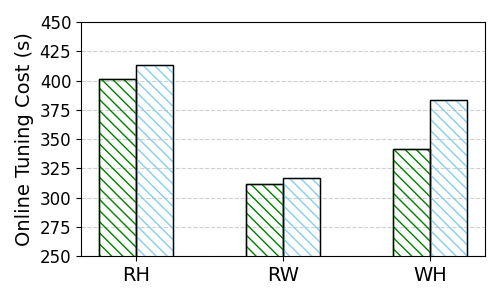}
		\subcaption{Online tuning cost}\label{fig:action guide cost} 
	\end{minipage}
    \caption{The performance of DemoTuner with and without hint priority update across different YCSB workloads on PostgreSQL.}\label{fig:action guide ablation exp}
\end{figure}

During the fine-tuning phase of agent offline training, some tuning hints may present diminishing benefits along with the training  steps and even deviate from the tuning target. To validate the necessity of hint priority update technique in the HA-DDPGfD algorithm, we trained DemoTuner with and without this technique and then compare their performance of online knobs tuning across three different workload types on PostgreSQL.\ As shown in Figure~\ref{fig:action guide ablation exp}, DemoTuner can always achieve a better performance gain while consuming a less online tuning cost with the priority update of tuning hints. For instance, under the Write-Heavy workload, regarding tuning hints' priorities as static constants loses 6.64\% performance gain even spending 12.12\% more online tuning cost than DemoTuner. By employing the hint priority update technique,  DemoTuner  can adaptively adjust the perturbation direction of actions to align with both domain knowledge and the current runtime environment,
thereby achieving a better agent for the online knobs tuning task.


\setlength{\abovecaptionskip}{0.4 cm}
\setlength{\belowcaptionskip}{-0.3 cm}

\setlength{\abovecaptionskip}{0.4 cm}
\setlength{\belowcaptionskip}{-0.4 cm}
    \begin{figure}[t]
    \centering
    \includegraphics[width=0.7\linewidth]{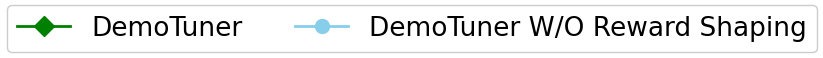}
    \includegraphics[width=0.48\textwidth]{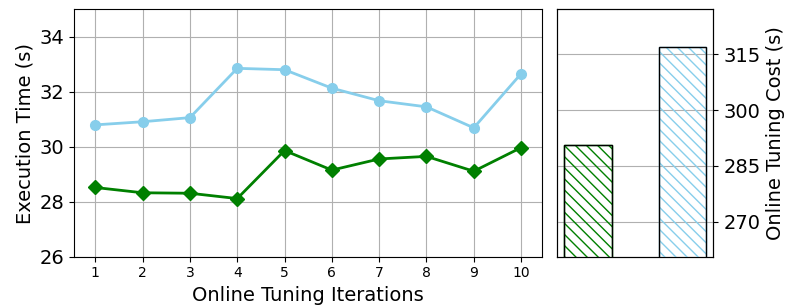} 
    \caption{The performance of DemoTuner with and without hint-guided reward shaping under RW workload on PostgreSQL.}
    \label{fig:reward-shaping comparison}
\end{figure}

\subsubsection{Effectiveness of Hint-Guided Reward Shaping}
To evaluate the effectiveness of the proposed hint-guided reward shaping technique, we trained DemoTuner with and without this technique and then evaluate their online tuning performance under the Read-Write workload on PostgreSQL.\ Figure~\ref{fig:reward-shaping comparison} displays the execution time for each configuration generated during the 10 online tuning iterations as well as the total online tuning cost. With hint-guided reward shaping, DemoTuner is able to penalize the action's reward by a specific range according to the number of current matched tuning hints the action disobeys. In this way, DemoTuner can incorporate the domain knowledge contained in the tuning hints to provide informative reward signals, so as to facilitate the convergence rate of offline training. Consequently, DemoTuner with reward shaping achieves an additional 8.39\% execution time reduction by the found best configuration while saving as much as 8.35\% total online tuning cost.

\subsection{ Discussions of Hyperparameter Settings}\label{subsec:hyper_param_in_DEMTune}

\subsubsection{Enhancement Ratio of Demonstration Sampling Probability $\lambda_{2}$}\label{sec:7-d-2}
During the agent fine-tuning phase of DemoTuner, both demonstrations from tuning hints and transitions from environmental feedback are stored in the experience replay buffer and appropriately utilized according to the proposed hpPER mechanism. In hpPER, the sampling probability of each experience is calculated  using the formula in Eq. (\ref{eq-3}), where an enhancement ratio $\lambda_2$ and the current priorities of tuning hints determine the enhanced probability for the demonstrations. Thus, the value of $\lambda_2$ is crucial for appropriately assigning sampling probabilities to each transition and demonstration. To determine the value of $\lambda_2$, we evaluate the performance of DemoTuner when the enhancement ratio is set to be different candidates. The experiments are conducted under the RW workload on PostgreSQL with 10 online tuning iterations. As shown in Figure \ref{transiton-importance}, when $\lambda_2$ is set to be 0, demonstrations and transitions stored in the experience replay buffer will be treated equally, which may lose opportunities of employing domain knowledge to further improve the agent training. On the other hand, a larger $\lambda_2$  may lead to an excessive dependence on the tuning hints and overlook the functions of transitions with practical feedback. Accordingly, we set $\lambda_{2}=1$ considering both the execution time reduction and the online tuning cost. 


\subsubsection{Control Coefficient of Reward Shaping Extent $\beta$}\label{subsubsec:reward_shaping_ratio q}
\begin{figure}[t]
    \centering
    \captionsetup[sub]{skip=-0.0cm}
	\begin{minipage}{0.49\linewidth}
		\centering
		\includegraphics[width=1\textwidth]{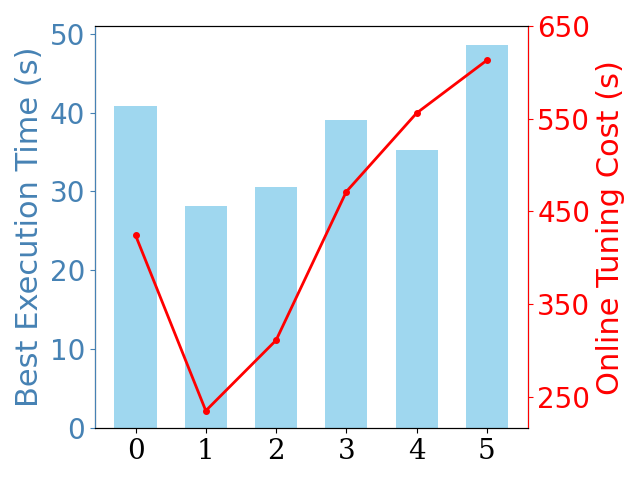}
		\subcaption{Enhancement Ratio $\lambda_2$}\label{transiton-importance} 
	\end{minipage}
	\begin{minipage}{0.49\linewidth}
		\centering
		\includegraphics[width=1\textwidth]{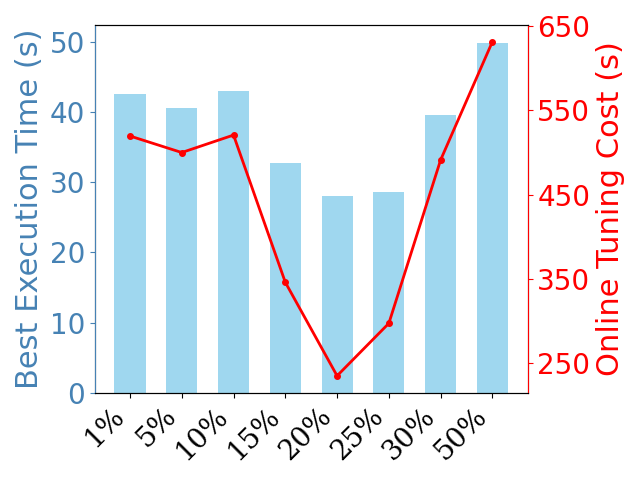}
		\subcaption{Control Coefficient $\beta$}\label{reward-shaping context} 
	\end{minipage}
    \caption{The performance of DemoTuner under different hyperparameter settings under the RW
workload on PostgreSQL.}\label{fig:hyperparameters}
\end{figure}

DemoTuner employs a hint-guided reward shaping technique to adjust an action's reward according to the tuning hints it goes against. Given that reward is a critical metric in RL, the extent of reward shaping $f_t$ calculated in Eq. (\ref{equ:reward-shaping}) significantly impacts the agent training process. Generally, the recommended extent for reward shaping is between 5\% and 20\% of the original reward. Therefore, we evaluate DemoTuner when the control coefficient $\beta$ of reward shaping extent is set to be 1\%, 5\%, 10\%, 15\%, 20\%, 25\% and 50\%, respectively. All experiments are conducted under the RW workload on PostgreSQL with 10 online tuning iterations. As shown in Figure \ref{reward-shaping context}, when $\beta$ is too small (e.g., below 10\%), the reward fails to effectively guide the training process and leads to a slow convergence rate. Conversely, when $\beta$ is too large (e.g., above 30\%), the training process will be unstable and also results in a slow convergence speed. Based on these results, we set $\beta$ at 20\%.

\subsection{Query-level Performance Improvement Analysis}\label{subsec:query-level}
\begin{figure}[t] 
    \centering 
    \includegraphics[width=0.48\textwidth]{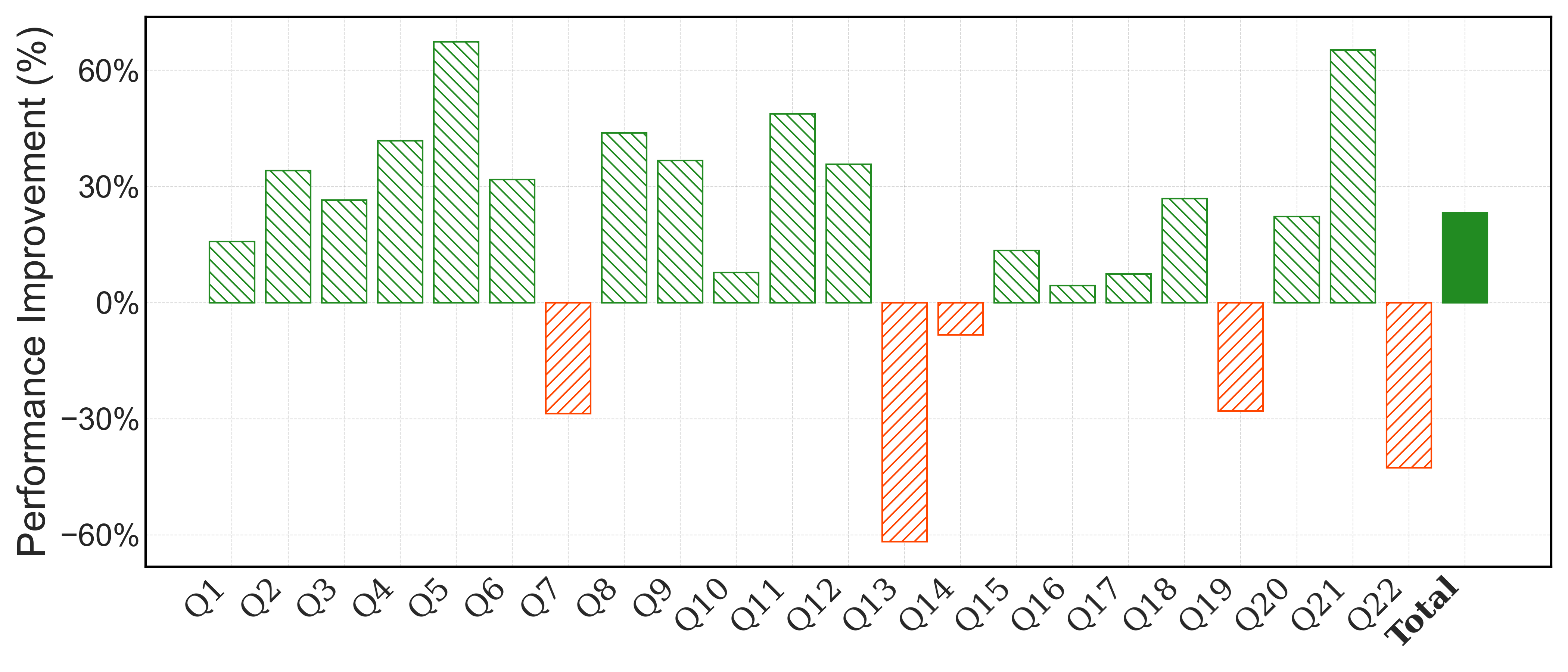} 
    \caption{Query-level performance improvement achieved by DemoTuner under TPC-H workload over default configuration.}\label{fig:tpch-query-performance-improvement} 
    \end{figure}

Like many previous workload-level DBMS knobs tuning studies, DemoTuner primarily targets  improving the overall performance of a DBMS under typical workloads. In contrast, query-level knobs tuning works~\cite{li2019qtune,xiu2025hint,chen2025aqetuner} focus on optimizing parameters that affect the execution of individual queries (often high-cost or critical ones). To provide a fine-grained view of DemoTuner’s workload-level performance improvement, we analyze its per-query effects over the default configuration on PostgreSQL with TPC-H (22 different queries in total). As shown in Figure~\ref{fig:tpch-query-performance-improvement}, DemoTuner improves the majority of queries (17/22), leading to the 23.23\% workload-level performance improvement over default. These gains primarily stem from the changes in query plan and execution procedure induced by knobs tuning such as \textit{Reduced Disk I/O}, \textit{Enhanced Memory Use}, \textit{Altered Parallelism}, etc. On the other hand, performance regressions of the other 5 queries reflect the inherent tension between  workload-level and query-level configuration tuning: knobs tuned for the overall workload-level performance might push the query optimizer toward choices that degrade specific queries. Extending DemoTuner to perform workload-level knobs tuning while guaranteeing per-query performance is a promising direction for future work. 


\setlength{\abovecaptionskip}{0.4 cm}
\setlength{\belowcaptionskip}{-0.3 cm}

\setlength{\abovecaptionskip}{-0.1 cm}
\setlength{\belowcaptionskip}{-0.4 cm}

\section{Conclusion}\label{sec-8}
In this paper, we propose DemoTuner, an efficient DBMS knobs tuning framework via a novel
LLM-assisted demonstration reinforcement
learning method. To comprehensively and accurately mine
tuning hints from documents, we design a structured CoT prompt to employ LLMs to conduct the condition-aware tuning hints extraction task. To effectively integrate the mined tuning hints
into RL agent training, we propose a hint-aware demonstration reinforcement learning algorithm HA-DDPGfD in DemoTuner. To accelerate the convergence speed, HA-DDPGfD first pre-trains the agent with the collected demonstrations and then fine-tunes the agent with the proposed hint priority-driven PER mechanism as well as a hint-guided reward shaping technique. Experimental evaluations conducted on MySQL and PostgreSQL across various workloads demonstrate the significant advantages of DemoTuner in both performance improvement and online tuning cost reduction over the baseline methods. Additionally, DemoTuner also exhibits superior adaptability to simulated application scenarios with unknown workloads. In the future, we will add support for more DBMSs in DemoTuner and study how to improve the effectiveness of LLM-driven DBMS knobs tuning methods.

\bibliographystyle{IEEEtran}
\bibliography{reference}


\end{document}